\documentclass[letterpaper]{article} % DO NOT CHANGE THIS
\usepackage{aaai23}  % DO NOT CHANGE THIS
\usepackage{times}  % DO NOT CHANGE THIS
\usepackage{helvet}  % DO NOT CHANGE THIS
\usepackage{courier}  % DO NOT CHANGE THIS
\usepackage[hyphens]{url}  % DO NOT CHANGE THIS
\usepackage{graphicx} % DO NOT CHANGE THIS
\usepackage{natbib}  % DO NOT CHANGE THIS AND DO NOT ADD ANY OPTIONS TO IT
\usepackage{caption} % DO NOT CHANGE THIS AND DO NOT ADD ANY OPTIONS TO IT
\usepackage{algorithm}
\usepackage{algorithmic}
\usepackage{multirow}
\usepackage{booktabs}
\usepackage{amsmath}
\usepackage{subfigure}
\usepackage{color}
\usepackage{newfloat}
\usepackage{listings}
\DeclareCaptionStyle{ruled}{labelfont=normalfont,labelsep=colon,strut=off} % DO NOT CHANGE THIS
\floatstyle{ruled}
\newfloat{listing}{tb}{lst}{}
\floatname{listing}{Listing}
\title{RenewNAT: Renewing Potential Translation for
Non-Autoregressive Transformer}
\author{
    Pei Guo\equalcontrib,
    Yisheng Xiao\equalcontrib,
    Juntao Li\thanks{Corresponding Author},
    Min Zhang
}
\affiliations{
    Institute of Artificial Intelligence, School of Computer Science and Technology, Soochow University\\
    \{pguolst,ysxiaoo\}@stu.suda.edu.cn, \{ljt,minzhang\}@suda.edu.cn
}

\usepackage{bibentry}
\begin{document}
\maketitle
\begin{abstract}
Non-autoregressive neural machine translation (NAT) models are proposed to accelerate the inference process while maintaining relatively high performance.
However, existing NAT models are difficult to achieve the desired efficiency-quality trade-off.
For one thing, fully NAT models with efficient inference perform inferior to their autoregressive counterparts.
 For another, iterative NAT models can, though, achieve comparable performance while diminishing the advantage of speed.
In this paper, we propose RenewNAT, a flexible framework with high efficiency and effectiveness, to incorporate the merits of fully and iterative NAT models.
RenewNAT first generates the potential translation results and then renews them in a single pass.
It can achieve significant performance improvements at the same expense as traditional NAT models (without introducing additional model parameters and decoding latency).
Experimental results on various translation benchmarks (e.g., \textbf{4} WMT) show that our framework consistently improves the performance of strong fully NAT methods (e.g., GLAT and DSLP) without additional speed overhead\footnote{\url{https://github.com/AllForward/RenewNAT}}.

% Specifically, RenewNAT firstly generates the potential translation results, then renews them in a single pass.
% Surprisingly, RenewNAT can achieve significant performance improvements with the same expense of traditional NAT models (without introducing additional model parameters and extra decoding latency).

% and even achieves comparable translation quality as AT models, while retaining the efficient inference.

% NAT目前存在的问题（fully iterative）
% 融合了两者的优势，
% 我们framework的特点和优势
% 效果

% Non-autoregressive neural machine translation (NAT) models are proposed to achieve inference acceleration while maintaining high translation quality. 
% However, existing NAT models are difficult to achieve a desired efficiency-quality trade-off. Either fully NAT models with efficient inference perform inferior to their autoregressive counterparts, or iterative NAT models achieve comparable performance while the speedup of inference declines. 
% In this paper, we propose RenewNAT, a novel framework, 

% Non-autoregressive neural machine translation (NAT) models are proposed to achieve inference acceleration while maintaining high performance. However, 
\end{abstract}
\section{Introduction}
\label{sec:intro}

Transformer-based models have been widely applied to neural machine translation (NMT)~\cite{dehghani2018universal,wu2019depth,wu2021r,takase2021rethinking,nguyen2020data,liang2022multi}. 
Despite the excellent performance of these methods, they normally adopt an autoregressive (AR) decoding paradigm in which the tokens are predicted one by one in a strict left-to-right order.
With such recurrent dependencies, the inference process is inevitably time-consuming, especially for long sentences.
% and with this attribute the inference may be quite time-consuming, especially for long sentences. 
To mitigate this problem, much attention has been paid to the fully non-autoregressive (NAT) models~\cite{xiao2022survey,gu2018non,ghazvininejad2019mask,qian2020glancing}, which can generate all tokens in parallel and thus accelerate the inference greatly.
However, this increase in inference efficiency is at the cost of the translation quality, and there exists a considerable performance gap between NAT models and their AT counterparts. 
Many works have explored the potential reasons and attributed the quality decline in NAT models to \textit{`the failure of capturing the target side dependency.'}~\cite{gu2020fully,xiao2022survey}. 
More specifically, fully NAT models learn to perform prediction without the internal dependency of the target sentence as a result of parallel decoding, unlike the AT models where the $t$-th token has previous $t$-$1$ contextual tokens' information to help its generation.

% \textcolor{red}{More specifically, fully NAT models learn to do prediction only conditioned on the source sentence as a result of parallel decoding, and this independent conditional assumption prevents the model learning internal dependency of target sentence, further leads to a performance gap.}
% NAT models generate the target tokens without the target-side information as a result of parallel decoding,
% where in AT models each token can get history prediction tokens as target contextual information to help prediction.

Significant efforts have been made to address the above-mentioned obstacles for fully NAT models from different aspects in recent years, e.g., adopting knowledge distillation~\cite{gu2018non,zhou2019understanding}, introducing latent variables~\cite{ran2021guiding,liu2021enriching}, training with better criterion~\cite{ghazvininejad2020aligned,du2021order} or learning strategies~\cite{qian2020glancing,bao2022glat,huang2022non}. \citeauthor{gu2020fully} also combine these tricks for better performance. 
However, there still exists a performance gap with their strong autoregressive counterparts, indicating that the translation results are still relatively not reliable. 
Specifically, these models only introduce effective tricks in the training process and still lack explicit target side information to guide translation during inference.
Thus, each target token is still predicted independently, and the well-known multi-modality problem~\cite{gu2018non} still hurts the translation performance seriously.
% Notice that one-step decoding prevents the model doing prediction depended on any target tokens and capturing the target side dependency.

% Specifically, the well-known multi-modality problem~\cite{gu2018non} 
% Although these works have improved fully NAT performance, they don't fundamentally solve the problem of lacking the target side dependency. To better capture the target-side information, generating potential results to refine is an important way. 
% So iterative NAT models are proposed by making use of this way.
% Specifically, since these fully NAT models discard the dependency of the target tokens, they predict each target token independently. Then they may choose fragments of different reasonable translation, leading to incomplete translation or repetitive translation, called multi-modality problem~\cite{gu2018non}. 
% Obviously, 
% Obviously, these models can not directly provide target tokens to help capture the target-side dependency 
% these models can not directly help the fully NAT models to capture the target side dependency and solve the multi-modality problem at the root.
% % Once the target tokens are generated, there is no chance for these models to do further operations depended on these target tokens. But it seems that fully NAT models require this critically.
% these methods can not  and the traditional . In the other word, 

Another representative method is introducing iterative refinements for NAT models during inference~\cite{ghazvininejad2019mask,huang2022improving}, called iterative NAT models.
Iterative NAT models adopt multiple decoding steps and keep the non-autoregressive property in each step, where the tokens generated from the last step are fed into the model for refinements. 
Obviously, the result generated from the previous iteration can provide target-side information to help predict and do refinements in the next iteration. 
Compared with fully NAT models, iterative NAT models have achieved more exciting performance, even outperforming the vanilla Transformer.  
However, due to multiple decoding steps, the advantage of decoding speed will diminish, especially in some specific scenes.

\begin{figure*}[htbp]
\centering
\includegraphics[scale=0.6]{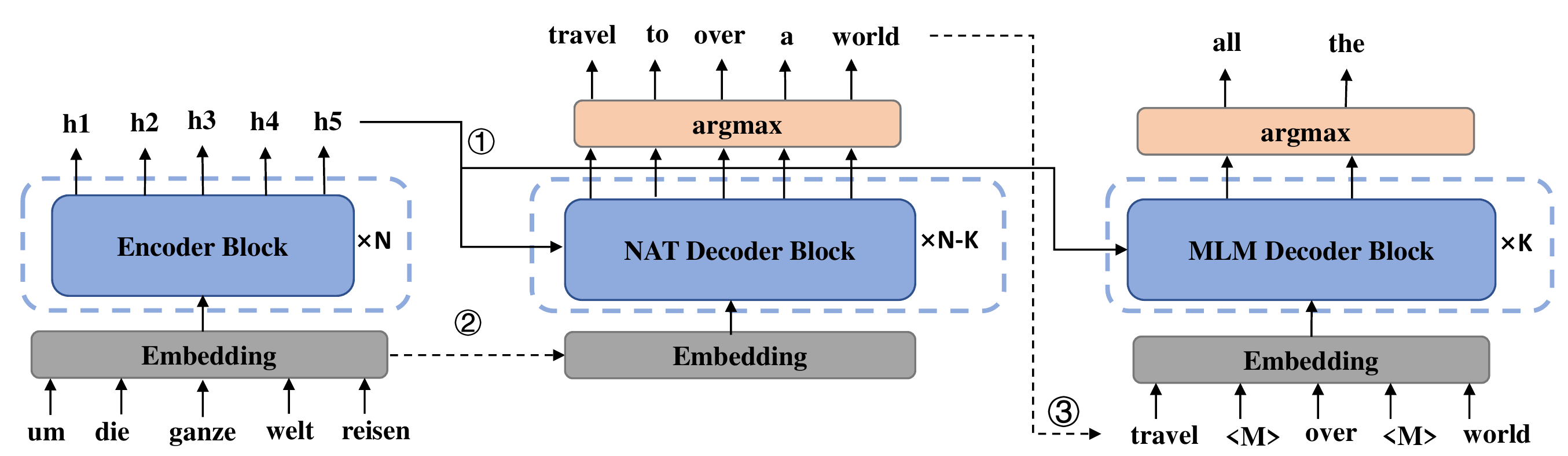}
\caption{The architecture of our RenewNAT. Different from the traditional fully NAT models, RenewNAT splits the decoder into two sub-modules. The first one aims to predict the potential translation, which can be viewed as a noisy version of the final translation. The input to it is the  embeddings of the source sentence with \textit{soft copy} (No.2 dashed arrow). The second one tries to renew the potential translation to a credible one by masking the incorrect tokens. The input to it
is the embeddings of potential translation generated by the first sub-module (No.3 dashed arrow).
The No.1 solid arrow denotes the encoder-decoder attention.}
\label{fig:RenewNAT_arch}
\end{figure*}

As a result, how to effectively combine the merits of the above-mentioned two NAT paradigms needs further exploration, e.g., to refine the inferior translation results with the help of useful target-side information without decreasing the inference speed.
In this work, we propose RenewNAT, a simple yet effective framework for non-autoregressive translation, which first generates the potential translation and then renews them to a credible one. 
Concretely, as shown in Figure~\ref{fig:RenewNAT_arch}, the decoder is split into two sub-modules according to the specific function, where the first aims to generate a potential translation containing useful target side information, while the second attempts to capture the target side dependency by deriving some valuable contextual information from the potential translation. 
However, as the potential translation can be viewed as a noisy version of the final translation, it inevitably contains errors.
These errors might be cascaded and amplified through the second sub-module, preventing it from capturing the correct target tokens dependency by directly learning from the potential translation.
To best derive valuable information from it, it is critical to discard the incorrect tokens in the potential translation, 
Motivated by the masked language model (MLM)~\cite{devlin2018bert,ghazvininejad2019mask}, we mask the incorrect tokens in the training of the second sub-module to keep the correctness of dependent target tokens.
Compared with latent variable NAT models, which also spilt the model into two sub-modules and the first one is adopted to predict a latent variable sequence to help prediction during inference, our RenewNAT aims at providing more specific target side information to guide prediction.
Moreover, latent variable NAT models need extra tools to provide latent variable information during training and this will increase the difficulty of training. 
In contrast, the first sub-module in our RenewNAT can be directly learned with ground truth tokens and this does not bring any additional expense. 

% It seems our RenewNAT is similar to the latent variable NAT models, which first predict a latent variable sequence containing a chunk of words or some other prompt information to help capture the target side dependency of tokens, such as target linguistic information~\cite{shu2020latent,ma2019flowseq}, alignments information~\cite{gu2018non,song2021alignart}, position information~\cite{ran2021guiding,bao2019non} and syntactic information~\cite{bao2021non,liu2021enriching}. However, potential translation is truly different from these latent variables and more directly to learn since it can be viewed as a noisy version of final translation. 

Our RenewNAT is a generic framework in which various widely-used methods can be adopted to improve the quality of potential translation and provide more useful target side information for the second sub-module. In experiments, we evaluate our RenewNAT on IWSLT'14 English$\rightarrow$German, WMT’14 English$\leftrightarrow$German, WMT’16 English$\leftrightarrow$Romanian and adopt a few effective and representative methods to improve the potential translation, such as GLAT~\cite{qian2020glancing} and DSLP~\cite{huang2022non}. 
In a nutshell, the major contributions of our paper are as follows. 
\begin{itemize}
    % \item We point out that the existing fully NAT models only introduce effective tricks during training and there is no specific target information to guide prediction during inference. Thus the critical problem that \textit{`the failure of capturing the target side dependency'} can not be effectively alleviated.
    \item We propose RenewNAT, which combines the merits of fully NAT and iterative NAT models. By splitting the model into two sub-modules, where the first one can provide useful target-side information to help prediction for the second one, we can alleviate the above-mentioned problem and meanwhile maintain the speed advantage.
    \item Experimental results on multiple machine translation benchmarks demonstrate that RenewNAT consistently improves the base model without the expensive cost of inference speed. Compared with the vanilla NAT model, RenewNAT achieves significant improvements by up to 2.0$\uparrow$ of BLEU score on average.
Compared with strong fully NAT models, e.g. GLAT~\cite{qian2020glancing} and DSLP~\cite{huang2022non}, RenewNAT also shows its effectiveness by up to 0.62$\uparrow$ of BLEU on GLAT and 0.3$\uparrow$ of BLEU on DSLP, maintaining about 13 times speedup.
\end{itemize}

\section{Related Work}
Non-autoregressive models have attracted much attention in the field of sequence generation, such as machine translation~\cite{gu2018non}, speech recognition~\cite{higuchi2021comparative}, text summarization~\cite{liu2022learning}, grammatical error correction~\cite{straka2021character}, dialogue~\cite{han2020non} and etc. Compared with their auto-regressive counterparts, which generate the target tokens from left to right, non-autoregressive models predict the target sequence in parallel and accelerate the speed of inference greatly, but the generation quality declines.
In this paper, we mainly focus on non-autoregressive translation (NAT)~\cite{gu2018non,guo2019non,ghazvininejad2020aligned,gu2020fully,ran2021guiding,ding2021progressive,qian2020glancing,huang2021non,zhan2022non,bao2022glat}, where much progress has been made to improve the performance in recent years. 

The NAT model is first proposed in \citet{gu2018non}, they adopt a fertility predictor to decide the number of the source token that will be aligned to and then help prediction. However, the translation performance is much lower than the vanilla Transformer because of the parallel decoding.
They notice that the NAT model can not make predictions well without capturing the target side dependency, further leading to the serious multi-modality problem. Many methods have been proposed to alleviate this in recent years. \citet{guo2019non} propose phrase-table lookup
and embedding mapping methods to enhance the decoder inputs with some target side information.
\citet{sun2019fast} utilizes conditional random fields to model the target positional contexts. \citet{chan2020imputer} tries to model the latent alignments to help prediction. 
\citet{ghazvininejad2020aligned} and \citet{du2021order} introduce better criterion 
for NAT models to learn. Besides, utilizing latent variables as part of the model is also a
popular method, such as the alignments between source and target tokens~\cite{song2021alignart}, positional information~\cite{bao2019non,ran2021guiding}, syntactic labels~\cite{akoury2019syntactically,liu2021enriching}.
At the same time, to best find the trade-off between translation speed and quality, much progress has also been made for iterative NAT models. Instead of generating all target
tokens in one pass, they learn the conditional distribution over partially observed generated tokens and adopt multiple steps to refine the previous results. There exist many specific ways for refinements, such as 
heuristic denoising~\cite{lee2018deterministic,savinov2021step}
insertion and deletion~\cite{stern2019insertion,gu2019levenshtein}
masking and recovering~\cite{ghazvininejad2019mask,kasai2020parallel} and so on.
However, these methods improve translation accuracy at the expense of speedup.
Recently, motivated by curriculum learning, \citet{qian2020glancing} introduces a glancing sampling strategy to dynamically select some positions based on the performance of the model, this significantly improves the performance of fully NAT models. Based on their simple method and excellent performance, many works further improve the performance. 
\citet{gu2020fully} attributes the key to fully NAT models to adopting dependency reduction in the learning space of output tokens. They combine several useful skills during training and finally achieve several SOTA results. 
\citet{huang2022non} introduce deep supervision and additional layer-wise
prediction (DSLP) for each decoder layer and \citet{zhan2022non} propose a general approach
to enhance the target dependency within the NAT decoder from decoder input and decoder self-attention, setting new SOTA results for fully NAT models. 
However, many methods for fully NAT models just introduce effective tricks in the training process, during inference, they are still unable to get any specific target side tokens to help do prediction, e.g., latent variables only model the latent information and this is different with the specific tokens, GLAT~\cite{qian2020glancing} and DSLP~\cite{huang2022non} keep the decoding process unchanged which prevents the model predicting tokens conditional on target side tokens. There is still room for improvement by modifying the traditional decoding process for fully NAT models.

\section{Methodology}
In this section, we first give a brief description of fully NAT models and iterative NAT models. Then, we present our RenewNAT framework in detail, including its training and inference process with some effective schemes.   

% we first describe fully NAT model. Then we propose our unified framework based on fully NAT model named RenewNAT which is able to get history prediction tokens as target contextual information to further predict with single-pass parallel generation. What's more, we present the process of training and inference in detail.

\subsection{Non-Autoregressive Transformer}
Vanilla NAT models are built upon the Transformer architecture~\cite{vaswani2017attention} which adopts the encoder-decoder network. 
Given a dataset $D=\{(X, Y)_i\}_{i=1}^N$, where $(X, Y)_i$ is one of the paired sentence data, and $N$ is the size of the dataset. 
$X=\{x_1,x_2,...,x_{T_X}\}$ is the source sentence to be translated from source language $\mathcal{X}$ and $Y = \{y_1, y_2, ..., y_{T_Y}\}$ is the ground-truth sentence from target language $\mathcal{Y}$, where $T_X$ and $T_Y$ are the length of source and target sequence.
The encoder is adopted to obtain the context information in $X$ by mapping $X$ to a feature vector $H_{enc}$, and the decoder adopts $H_{enc}$ and the decoder input $\hat{Y}$ to generate the translation results $Y$. 
The goal of NAT models is to estimate the unknown conditional distribution $P(Y|X; \theta)$ by learning a mapping function $f(\cdot)$ from the source sentence to the target sentence, where $\theta$ denotes the parameter set of a NAT model. 
During training, fully NAT models use the conditional independent factorization for prediction, and the objective is to maximize:
\begin{equation}
% \[
\mathcal{L}_{\text{NAT}}=\sum_{t=1}^{T} \log P(y_t|X;\theta), \label{Fully}
% \]
\end{equation}
where $T$ is the length of the target sentence. During training, $T$ is the same as the length of the ground-truth target sentence $T_Y$, while in inference, $T$ is usually predicted by a length prediction module $P_L$.
Compared with their AT counterparts, it is obvious that the conditional tokens $y_{<t}$ are removed for NAT models. 
Hence, we can translate in parallel without auto-regressive dependencies, and improve the inference speed greatly.

For iterative NAT models, they also adopt the conditional independence assumption identically, but they try to rebuild the dependency of the target tokens in an iterative fashion. During inference, they keep the non-autoregressive property in every iteration step and refine the translation results during different iteration steps. Specifically, in the first iteration, only $X$ is fed into the model, which is the same as NAT models. After that, each iteration takes the translation generated from the last iteration as context for refinement to decode the translation. 
The training goal of the iterative-based NAT models is to maximize:
\begin{equation}
% \[
\mathcal{L}_{\text{Iter}}=\sum_{y_t \in Y_{tgt}} \log P(y_t|\hat{Y},X;\theta), \label{Iter}
% \]
\end{equation}
where $\hat{Y}$ indicates the translation result of the last iteration,
% with some specific operations, 
and $Y_{tgt}$ is the target of current iteration.
Taken CMLM~\cite{ghazvininejad2019mask} as an example, it explicitly learns the conditional distribution over partially observed reference tokens, and then during inference, several tokens with low confidence will be masked and predicted again in the next iteration based on the tokens unmasked.
% while the tokens unmasked can provide useful target side information to guide prediction.

% Based on fully NAT, the iterative generation approaches use the conditional independence assumption identically, but they rebuild the dependency of the target tokens in an iterative fashion. For example, CMLM (Ghazvininejad et al., 2019) adopts the random masking scheme to choose a set of tokens from target sentences and replaces them with the [MASK] token, then provides the processed sentences as the decoder input. The training objective of CMLM is to predict the masked words given the source sentence $X$ and the processed sentences, it could be described formally by follow: 
% \begin{equation}
% % \[
% \mathcal{L}_{\text{CMLM}}=\sum_{y_t \in f_{ran}(Y))} \log P(y_t|\beta(Y, f_{ran}(Y)),X). \label{CMLM}
% % \]
% \end{equation}
% where $f_{ran}(Y)$ is a set of tokens randomly chosen from \textit{Y}, and $\beta(Y, f_{ran}(Y)$ denotes that the chosen tokens are replaced with the [MASK] token. 

% During inference, CMLM is able to build the dependency of the target tokens by refining the words generated in the previous iteration. In a word, the iterative NAT models greatly improve the performance but lose the advantage of inference speed.

\subsection{RenewNAT Framework}
In order to effectively alleviate the problem mentioned above, we hope our proposed model is able to do prediction conditioned on specific target tokens during inference while maintaining the speed advantage without iterative operation.
We propose RenewNAT, a simple yet effective framework. As shown in Figure~\ref{fig:RenewNAT_arch}, RenewNAT splits the traditional NAT decoder into two sub-modules, called NAT sub-module and MLM sub-module. 
The NAT sub-module aims to predict the potential translation to provide necessary target tokens while the MLM sub-module can derive some valuable contextual information and then renew the potential translation to a credible one.
Next, we will introduce these two sub-modules in detail.
% to make the model parameter number comparable with the baseline model, RenewNAT divides the decoder into two categories and use \textit{K} and \textit{N - K} decoder blocks for fully NAT decoder module and MLM module respectively. Fully NAT decoder module is used to generate the potential translation results, and MLM module is adopted to renew the potential translation results. Next, .

\subsubsection{NAT Sub-module}
Formally, given the source sentence $X$, the encoder takes $X$ as input and captures the contextual information by $N$ transformer encoder layers, denoted by: 
\begin{equation}
% \[
E_X=\text{Enc}_{1:N}(X)= \text{Layer}_{1:N}^{enc}(emb(X)), \label{enc}
% \]
\end{equation}
where $\text{Enc}_{1:N}$ denotes the encoder module, $\text{Layer}_{1:N}^{enc}$ denotes the first to $N$th encoder layer, $emb(\cdot)$ is the embedding function used to map the discrete words into vector representations and $E_X$ is the encoder’s hidden representation of the last layer, and  
the input to our NAT sub-module $H= \{h_1,h_2,...,h_T\}$ is copied from $emb(X)$ with \textit{uniform copy} or \textit{soft copy}, as:
\begin{equation}
H = f(emb(X)),
\end{equation}
where $f(\cdot)$ is the copy function. 
Then NAT sub-module will predict the potential translation with a softmax layer based on the hidden states of the last layer, denoted by:
\begin{equation}
% \[
E_{Y_{pot}}=\text{Dec}_{\text{1:\textit{N-K}}}(H,E_X) = \text{Layer}_{\textit{1:N-K}}^{dec}(H, E_X) \label{natdec}
% \]
\end{equation}
\begin{equation}
P(Y_{pot}| E_{Y_{pot}} ) = \text{softmax}(W\cdot E_{Y_{pot}}), \label{P:1:N-K}
\end{equation}
where $E_{Y_{pot}}$ is the hidden states of the last layer of NAT sub-module, $\text{Dec}_{1:N-k}$ denotes the NAT sub-module, $\text{Layer}_{1:N-k}^{dec}$ denotes the first to $N-k$th decoder layer, $W$ is a trainable matrix, $Y_{pot}$ is the potential translation results of our NAT sub-module. 
Since the potential translation can be viewed as a noisy version of the final translation, we can simply train our NAT sub-module with ground-truth tokens, so the training process is the same as fully NAT models in Equation~\ref{Fully}, we denote the training loss as $\mathcal{L}_{\text{pot}}$ here:
\begin{equation}
\mathcal{L}_{\text{pot}}=\sum_{t=1}^{T} \log P(y_t|X;\theta). \label{potloss}
\end{equation}

% $Layer_{1:N}^e$ represent the first to Nth encoder layer, and $emb$ is an embedding function used to map the discrete words into vector representations.

% In our RenewNAT, fully NAT decoder module contains \textit{N-K} transformer decoder layers, which is used to generate the potential translation results. Thus, fully NAT decoder module could be represented by 
% \begin{equation}
% % \[
% Dec_{\text{1:\textit{N-K}}} = Layer_{1:N-K}^d(H, Enc_{1:N}). \label{dec}
% % \]
% \end{equation}
% where $Layer_{1:N-K}^d$ represent the first to N-Kth decoder layer, and \textit{H} is copied from the encoder output such as using soft copy. Finally, fully NAT decoder module uses a softmax layer to generate the potential words based on $Dec_{1:N-K}$: 
% \begin{equation}
% % \[
% P(Y|Dec_{1:N-K}) = softmax(W\cdot Dec_{1:N-K}). \label{P:1:N-K}
% % \]
% \end{equation}
% where \textit{W} is a trainable matrix. Thus the training objective $L_{fully}$ is the same as $L_{NAT}$

\subsubsection{MLM Sub-module}
After NAT sub-module predicts the potential translation $T_{pot}$, MLM sub-module aims to 
derive some valuable contextual information from it and then renew it to a credible one during inference. 
We do not change the internal structure of MLM sub-module layers, but to match the inference process mentioned above specifically, we change the training scheme to urge it to rebuild the dependency of the target tokens. Fortunately, we find the training scheme in CMLM~\cite{ghazvininejad2019mask} with some modification can satisfy our assumption. 
While the tokens masked can be viewed as incorrect tokens in potential translation and the tokens unmasked can be viewed as valuable contextual information. 
Formally, given the target sentence $Y$, MLM sub-module adopts the uniform masking scheme to choose a set of target tokens and replace them with [MASK] token. As a result, $Y$ is divided into $Y_{mask}$ and $Y_{obs}$, denoted as $Y'$, then serves as the input of MLM sub-module. MLM sub-module also utilizes the encoder output $E_X$ into an encoder-decoder attention module to obtain source information. We can describe the modeling process as follow:
\begin{equation}
\begin{aligned}
E_Y & =\text{Dec}_{\textit{N-K+1:N}}(Y', E_X) \\ & = \text{Layer}_{\textit{N-K+1:N}}^{dec}(emb(Y'), E_X) \label{D:N-K:N}
% \]
\end{aligned}
\end{equation}
\begin{equation}
P(Y_{mlm}| E_{Y} ) = \text{softmax}(W'\cdot E_{Y}), \label{P:N-k+1:k}
\end{equation}
where $E_{Y}$ is the hidden states of the last layer of MLM sub-module, $\text{Dec}_{\textit{N-K+1:N}}$ denotes MLM sub-module, $\text{Layer}_{\textit{N-K+1:N}}^{dec}$ denotes the ($N-K+1$)th to $N$th decoder layer, $W'$ is a trainable matrix, $Y_{mlm}$ is the outputs of MLM module, notice that only the tokens in masked position are valuable. Such during training, we compute the loss of masked tokens only, denoted by $\mathcal{L}_{\text{mlm}}$, which can be computed as:
\begin{equation}
\mathcal{L}_{\text{mlm}}= -\sum_{y_t \in Y_{mask}} \log P(y_t|Y_{obs},X;\theta). \label{mlmloss}
\end{equation}

% used to refine those unconfidently-predicted target words via confidence. Thanks to the potential translation result contain the target information, better performance can be obtained by updating. As we can see, MLM module is able to receive the target information from fully NAT decoder module and provide the global context-dependence information of  target to renew the sentences without creating a gap between training and inference. 

% Thus, MLM module adopts the random masking scheme to choose a set of target tokens and replace them with [MASK] token, then provide them as the input. What's more, it also utilize the encoder output $Enc_{1:N}$ as an encoder–decoder attention to obtain input information. Formally, we can described the procession as follow:
% \begin{equation}
% % \[
% Dec_{\textit{N-K:N}} = Layer_{N-K:N}^d(\beta(Y, f_{ran}(Y), Enc_{1:N}). \label{D:N-K:N}
% % \]
% \end{equation}
% where $Layer_{N-K:N}^d$ represent the \textit{N-Kth} to \textit{Nth} decoder layer. Finally, MLM module also uses a softmax layer to generate the potential words based on $Dec_{N-K:N}$: 
% \begin{equation}
% % \[
% P(Y|Dec_{N-K:N}) = softmax(W\cdot Dec_{N-K:N}). \label{P:N-K:K}
% % \]
% \end{equation}
% The training objective $L_{mlm}$ is the same as $L_{CMLM}$. Therefore, the final loss that we use in RenewNAT is a combination of $L_{fully}$ and $L_{mlm}$:
% \begin{equation}
% % \[
% L = L_{fully} + L_{mlm} \label{NAT+MLM}
% % \]
% \end{equation}
\begin{table*}[!htb]
\centering
\small
\renewcommand\arraystretch{0.95}
\tabcolsep 4pt
 \scalebox{0.9}{
\begin{tabular}{lcl|c|cccc|c}
\toprule
\multicolumn{3}{c|}{\multirow{2}{*}{Models}} & \multirow{2}{*}{\qquad$I_{\text{dec}}$\qquad} & \multicolumn{2}{c}{WMT14} &\multicolumn{2}{c|}{WMT16} &\multirow{2}{*}{\qquad Speedup\qquad} \\
\multicolumn{3}{c|}{} &  & EN$\rightarrow$DE & DE$\rightarrow$EN& EN$\rightarrow$RO & RO$\rightarrow$EN &        \\
\midrule
\multicolumn{2}{c}{\multirow{3}{*}{AT Models}}  
  &Transformer~\cite{vaswani2017attention} & T & 27.30 & / & / & / &  / \\
  & & Transformer~\cite{qian2020glancing}   & T & 27.48 & 31.27 & 33.70 & 34.05 & 1.0$\times$ \\ 
  & & Transformer~(ours)~w/ KD & T & 28.47 & 31.95 & 33.70& 33.75 & 1.0$\times$\\
 
\midrule
 \multicolumn{2}{c}{\multirow{9}{*}{Iterative NAT}}& NAT-IR~\cite{lee2018deterministic}   & 10   & 21.61 & 25.48 & 29.32 & 30.19 & 1.5$\times$ \\
& & LaNMT~\cite{shu2020latent}  & 4     & 26.30 & /     & /     & 29.10 & 5.7$\times$  \\
&& LevT~\cite{gu2019levenshtein}  & 6+     & 27.27 & /     & /     & 33.26 & 4.0$\times$  \\
 %& Mask-Predict$_\text{small}$~($I_{dec}$=10)&25.51& 29.47& 31.65 & 32.27 & / \\
%  & Mask-Predict & 1 & 18.05 & 21.83 & 27.32 & 28.20 &  /           \\
&& Mask-Predict~\cite{ghazvininejad2019mask} & 10 &27.03& 30.53 & 33.08 & 33.31 & 1.7$\times$    \\
&& JM-NAT~\cite{guo2020jointly} & 10 & 27.69 & 32.24 & 33.52 & 33.72 & /\\
&& DisCO~\cite{kasai2020parallel} & Adv & 27.34 & 31.31 & 33.22 & 33.25 & 3.5$\times$\\   
&& Multi-Task NAT~\cite{hao2021multi} & 10 & 27.98  & 31.27 & 33.80 & 33.60 & 1.7$\times$\\  
&& RewriteNAT~\cite{geng2021learning} & Adv & 27.83 & 31.52 & 33.63  & 34.09 & /\\
&& CMLMC~\cite{huang2022improving} & 10 & 28.37 & 31.47 & 34.57 & 34.13 & / \\
 \midrule
\multirow{21}{*}{\ \ Fully NAT\ \ } &
 %& NAT-IR              & 13.91 & 16.77 & 24.45  & 25.73 &  9.0$\times$
 %\\
%  & NAT-REG              & 1 & 20.65 & 24.77 & /     & /     & / \\%27.6$\times$ \\
 &  NAT-FT~\cite{gu2018non} & 1 & 17.69 & 21.47 & 27.29 & 29.06 & 15.6$\times$ \\
 & & Mask-Predict~\cite{ghazvininejad2019mask} & 1 & 18.05 & 21.83 & 27.32 & 28.20 &  /           \\
  & & imit-NAT~\cite{wei2019imitation} & 1 & 22.44 & 25.67 & 28.61 & 28.90 & 18.6$\times$ \\
%  & & NAT-HINT~\cite{li2019hint}  & 1 & 21.11 & 25.24 & /     & /     & / \\%30.2$\times$ \\
 %  & Flowseq              & 1 & 21.45 & 26.16 & 29.34 & 30.44 &  /           \\
 & & Flowseq~\cite{ma2019flowseq} & 1 & 23.72 & 28.39 & 29.73 & 30.72 & 1.1$\times$\\
 & & CTC~\cite{saharia2020non} & 1 & 25.70 & 28.10 & 32.20 & 31.60 & 18.6 $\times$ \\
 & & Imputer~\cite{saharia2020non} & 1 &  25.80 &28.40 &32.30 &31.70 & 18.6 $\times$ \\
 & & ReorderNAT~\cite{ran2021guiding}        & 1 & 22.79 & 27.28 & 29.30 & 29.50 & 16.1 $\times$ \\
 & & SNAT~\cite{liu2021enriching} & 1& 24.64 &28.42 &32.87 &32.21 &22.6$\times$  \\
 & & AlignNAT~\cite{song2021alignart} & 1 & 26.40 &30.40 &32.50 &33.10 &11.3$\times$\\
 & & Fully-NAT~\cite{gu2020fully} & 1 & 26.51 & 30.46 & \textbf{33.41} & \textbf{34.07} & 16.5$\times$ \\
 & & OAXE-NAT~\cite{du2021order} & 1  &26.10 &30.20 &32.40 &33.30 &15.3$\times$  \\
 & & DAD~\cite{zhan2022non} & 1 & 26.43 &30.42 &33.07 &33.82 & 15.1$\times$ \\
%  & Mask-Predict$_\text{small}$~($I_{dec}$=1)  & 15.06 & 19.26 & 20.12 & 20.36 &  /           \\
%  & Mask-Predict$_\text{base}$ & 1 & 18.05 & 21.83 & 27.32 & 28.20 &  /           \\
 % & Imputer              & 1 & \textbf{25.8} & 28.4 & / & / & / \\
%  & NAT-base~(ours) & 1 & 20.36 & 24.81 & 28.47 & 29.43 &  15.3$\times$           \\
%  & \method~(ours)    & 1    & \textbf{25.21}& \textbf{29.84} & \textbf{31.19} & \textbf{32.04} & 15.3$\times$          \\
%\hline\
% \multicolumn{3}{c}{\multirow{9}{*}{Ours}}
\cmidrule{3-9}
% \multirow{9}{*}{\ \ Ours\ \ }
% \cmidrule{2-9}
% & \multirow{9}{*}{\textbf{Ours}}

& \multirow{3}{*}{}
 & $\dagger$Vanilla NAT  & 1 & 19.34 & 24.42 & 29.19 & 29.61  & 13.6$\times$ \\
 & & $\dagger$RenewNAT  & 1 & 22.38 & 27.16 & 31.01 & 32.30 & 13.1$\times$ \\
 & & $\dagger$RenewNAT w/ NPD (m=5) & 1 & 23.56 & 28.37 & 31.87 & 33.05 & 11.5$\times$\\
 \cmidrule{3-9}
 %\multicolumn{2}{c}{\multirow{5}{*}{\begin{tabular}[c]{@{}c@{}}Fully NAT \\ w/  NPD\end{tabular}}}
 & \multirow{5}{*}{} 
%  & NAT-FT~(m=10)         & 18.66 & 22.41 & 29.02 & 30.76 & 7.7$\times$  \\
 & $\dagger$GLAT  & 1 &  25.21 &  29.51 & 31.65  & 32.04 & 13.5$\times$  \\
 & & $\dagger$RenewNAT & 1 & 25.75 & 29.67 & 32.53 & 32.93 & 12.5$\times$ \\
 & & $\dagger$RenewNAT w/ NPD (m=5) & 1 & \textbf{26.54} & \textbf{30.55} & 32.86 & 33.35 & 10.7$\times$\\
%  & & GLAT + NPD~(m=7)  & 1 &  26.55 & 31.02 & 32.87 & 33.51 & 7.9$\times$  \\
%  & & \ \ w/ DSLP  & 1 & / & / & / & / & /$\times$ \\
%  & & \ \ w/ RenewNAT(\textbf{ours})  & 1 & / & / & / & / & /$\times$ \\
  \cmidrule{3-9}
 & & $\dagger$GLAT + DSLP & 1 &  25.69 & 29.90 & 32.24 & 33.03 & 12.6$\times$  \\
 & & $\dagger$RenewNAT  & 1 & 25.86 & 30.14 & 32.70 & 33.52 & 12.4$\times$ \\
 & & $\dagger$RenewNAT w/ NPD (m=5) & 1 & \textbf{26.65} & \textbf{30.65} & 33.02 & 33.74 & 11.2$\times$\\
 %  & Flowseq~(m=30)      & 1  & 23.48 & 28.40 & 31.75 & 32.49 & /           \\
%   \cmidrule{3-9}
%   & \multirow{3}{*}{}
%  & GLAT + CTC & 1 & - & - & - &-  & 15.7$\times$ \\
%  & & w/ RenewNAT(ours)  & 1 & / & / & - & / & /$\times$ \\
%  &\method~(NPD m=7, ours)  & 1   & \textbf{26.55} &\textbf{31.02}&\textbf{32.87}&\textbf{33.51} & 7.9$\times$  \\
  
%   & CTC~(ours)              & 1 & 25.52 & 28.73 & / & / & 14.6 $\times$ \\
%   & \method+CTC~(ours)              & 1 & 26.39 & 29.54 & / & / & 14.6 $\times$ \\
 \bottomrule
\end{tabular}
}
% }
\caption{Results on 4 WMT machine translation tasks. $\dagger$ denotes the results of our implementations. }% $N$ is the length of the output sequence.}
\label{tb.main_result}
\end{table*}

\begin{algorithm}[tb]
\caption{RenewNAT Training}
\label{alg:algorithm}
\textbf{Input}: training data $D = \{(X_i, Y_i)\}$, learning rate $\gamma$ \\
% correct threshold $\tau$ \\
\textbf{Output}: model parameters $\theta$ 

\begin{algorithmic}[1] %[1] enables line numbers
% \STATE Let $t=0$.
\STATE Initialize model with parameters $\theta$.
\WHILE{not convergent}
\FOR{$(X,Y)$ $\in$ $D$}
\STATE generate $\hat{Y}$ using NAT sub-module
% \STATE obtain current correct ratio $\eta$ $\gets$ $\hat{Y} \& Y$ 
% \IF {$\eta <$ $\tau$}
\STATE sample mask from $Y$: $Y_{mask}$ $\gets Y$ 
% \ELSE
% \STATE sample mask from $\hat{Y}$: $Y_{mask}$ $\gets \hat{Y}$ 
% \ENDIF
\STATE predict $Y_{mask}$ using MLM module
\STATE compute loss $\mathcal{L}$ $\gets$ $\mathcal{L}_{\text{pot}}+\mathcal{L}_{\text{mlm}}+\mathcal{L}_{\text{len}}$
\STATE update model parameters $\theta$ $\gets$ $\theta$ - $\gamma$ $\frac{\partial}{\partial \theta}$ $\mathcal{L}$
\ENDFOR
\ENDWHILE
\STATE \textbf{return} $\theta$
\end{algorithmic}
\end{algorithm}

\subsubsection{Training and Inference}
In the previous subsection, we introduced our RenewNAT framework and two decoder sub-modules. In order to describe the training procedure for RenewNAT clearly, Algorithm~\ref{alg:algorithm} is given to outline. 
During training, NAT sub-module and MLM sub-module are trained jointly. 
Besides, an extra module is trained to predict the target length, which is implemented as in previous works~\cite{qian2020glancing,huang2022non}. An additional [LENGTH] token is added to the source input, and the encoder output for the [LENGTH] token is used to predict the length. 
We denote the loss of the length prediction module as $\mathcal{L}_{\text{len}}$ here,
then the final training loss of our RenewNAT is $\mathcal{L}$:
\begin{equation}
\mathcal{L}=\mathcal{L}_{\text{pot}}+\mathcal{L}_{\text{mlm}}+\mathcal{L}_{\text{len}}. \label{loss}
\end{equation}
During inference, RenewNAT generates the translation with only a single pass. It first predicts the target length and then utilizes the NAT decoder sub-module to generate a potential translation result, then masks the incorrect tokens in potential translation and utilizes MLM decoder sub-module to renew them. Since it is not easy to distinguish the incorrect tokens in potential translation, we try many strategies and find that determining them by confidence is simple and effective. We set a threshold $\alpha$ to choose them, once the confidence is lower than $\alpha$, we regard them as incorrect tokens and can not provide valuable information. 
Moreover, the best $\alpha$ may depend on the base model that predicts the potential translation. 
We will discuss more in the following experiments.

\section{Experiment}

\subsection{Dataset}
We evaluate our RenewNAT on five widely used machine translation benchmarks, including WMT14 EN$\leftrightarrow$DE (4.5M pairs), WMT16 EN$\leftrightarrow$RO (610K pairs) and IWSLT14 DE$\rightarrow$EN (153K pairs). 
% These datasets are tokenized into subword units using byte-pair encoding (BPE) \cite{sennrich2016neural}. 
We follow the approach of~\citet{vaswani2017attention} to process WMT14 EN$\leftrightarrow$DE and adopt the approach to process data provided in~\citet{lee2018deterministic} for WMT16 EN$\leftrightarrow$RO. For IWSLT14 DE$\rightarrow$EN dataset, we follow the steps in \citet{guo2019non}. 
% We will compare the performance on raw and distill data at the same time.

\subsection{Knowledge Distillation}
Following the previous work~\cite{gu2018non,wang2019non,guo2019non}, we adopt the sequence-level knowledge distillation for all datasets. Firstly, a vanilla Transformer (transformer-large for WMT14 En$\leftrightarrow$DE, transformer-base for WMT16 En$\leftrightarrow$RO and IWSLT14 DE$\rightarrow$EN) is trained with raw data, and the results generated by the models on trained set are adopted as training data. Then we train RenewNAT with distilled data.

% we train a  transformer (transformer-large for WMT14 En $\leftrightarrow$ DE, transformer-base for WMT16 En $\leftrightarrow$ RO and IWSLT14 DE $\rightarrow$ EN) to generate the distill data which effectively alleviate multi-modal problems for NAT. Then we adopt the distill data for all baselines and RenewNAT.

\subsection{Evaluation Metrics}
To evaluate our model, we report BLEU scores \cite{papineni2002bleu} widely used in translation tasks to evaluate the performance.  
Speedup is measured by $L_{1}^{\text{GPU}}$ following the previous work~\cite{kasai2020deep,gu2020fully,helcl2022non}, specifically, we perform generation on the test set and set the batch-size as 1, then we compare the latency of RenewNAT with Vanilla Transformer on a single Nvidia A5000 card. We use all scripts from Fairseq~\cite{ott2019fairseq}.
% comet???

% (Papineni et al. 2002) widely used in translation tasks as evaluation metrics. Then we perform inference with batch-size = 1 to measure the inference latency on a single Nvidia A5000 GPU. We use all scripts from fairseq (Ott et al., 2019).

\subsection{Experimental Settings}
During training, we follow most of the hyperparameter settings in~\citet{gu2020fully,qian2020glancing}. 
For WMT datasets, we use base transformers configuration (6 layers per stack, 8 attention heads per layer, 512 model dimensions, 2048 hidden dimensions) and adapt the warm-up learning rate schedule~\cite{vaswani2017attention} with warming up to 5e-4 in 4k step. 
As IWSLT is a smaller dataset, we use a smaller Transformer model (6 layers per stack, 8 attention heads per layer, 512 model dimensions, 1024 hidden dimensions). 
We separately train the model with batches of 64k/8k tokens on WMT/IWSLT dataset and use Adam optimizer~\cite{kingma2014adam} $\beta$ with (0.9, 0.999) and (0.9, 0.98) for GLAT and Vanilla NAT.
We train all models for 300k steps and average the 5 best checkpoints chosen by validation BLEU scores as our final model for inference. 
It's noticed that we set \textit{K} as 2 because it could achieve the best performance evaluated in Ablation Study.
For AT models, we use a beam size of 5 to generate the results.
For RenewNAT, we apply noisy parallel decoding denoted as NPD~\cite{gu2018non} and set the length beam as 5.

% We separately train the model with batches of 64k/8k tokens and set the dropout rate to 0.1/0.2 for WMT/IWSLT datasets. And we use Adam optimizer (Kingma and Ba, 2014) with $\beta=(0.9, 0.999)$ . Employing label smoothing to 0.1 is provided. We train all models for 300k steps and create the final model by averaging the 5 best checkpoints chosen by validation BLEU scores (Papineni et al., 2002). During decoding, we use a beam size of b = 5 for autoregressive decoding, while length beam (Ghazvininejad et al., 2019) is applied to obtain length candidates.

\subsection{Baselines}
In order to evaluate the effectiveness of our framework, 
we apply our method on vanilla NAT model~\cite{gu2018non} and two strong NAT models, GLAT~\cite{qian2020glancing} and DSLP~\cite{huang2022non}, which significantly improve the performance of fully NAT models.
% and the latter one also achieves the state-of-the-art performance recently. 
For the auto-regressive baseline, we compare our method with the base Transformer trained with distilled data.
We also report the results of the recent fully NAT models and iterative NAT models for better comparison.
% In order to evaluate the effectiveness of our framework, we apply our methods on two fully NAT models: 1) Vanilla NAT (Gu et al., 2018) which is a highly representative NAT models. 2) GLAT (Qian et al., 2021), which is a highly competitive NAT models. At the same time, DSLP (Huang et al., 2021) is applied on these base models to form a strong baseline. With the help of these base models, we evaluate RenewNAT with the representative baselines.

\subsection{Main Results}
As shown in Table 1, we evaluate our RenewNAT with different base models.
% To best keep the speed advantage, we do not apply noisy parallel decoding which adopts multiple length candidate to do prediction and choose the best one.
% We also apply noisy parallel decoding and set the length candidate as 5.
Compared with vanilla NAT, RenewNAT gains significant improvements on each dataset, especially on WMT14 DE$\rightarrow$EN and WMT16 RO$\rightarrow$EN datasets (up to +2.2 and +2.6 BLEU). 
Besides, upon two stronger fully NAT models, GLAT and DSLP, RenewNAT still achieves about 0.5 BLEU improvement, while keeping the inference speed and model parameters unchanged.
The consistent improvements also verify the effectiveness of our approach and demonstrate that RenewNAT could be applied on various fully NAT models. 
Although the performance is still below strong iterative NAT models and their AT counterparts, its merit in decoding speed is significant.

% As shown in Table 1, we apply our methods to Vanilla NAT (Gu et al., 2018) and GLAT (Qian et al., 2021) on the WMT and IWSLT datasets. In order to compare our framework with the baselines fairly, we reproduce the results which are marked by * in Table~\ref{tb.main_result}. As we can see, RenewNAT achieves higher performance for two base NAT models and every translation task. RenewNAT consistently outperforms the base models by more than \textbf{0.5} BLEU score, which is able to verify the effectiveness of our approach and demonstrate RenewNAT could be applied on various fully NAT models. 

% At the same time, RenewNAT can achieve 15.4 speed up which is comparable with the base NAT models. When combining with the strong CTC model, RenewNAT also achieves better preformance. Compared with iterative methods, CTC w/ RenewNAT produces very competitive translation quality while being approximately 4 times faster in inference. (CTC + GLAT + RenewNAT) 

% All in all, it's shown that our framework provides target-side information dependency to NAT and achieves higher performance. At the same time, RenewNAT maintains the speed advantage of fully NAT and produces the competitive performance compared with iterative NAT models.

% All in all, RenewNAT maintains the speed advantage of fully NAT and produces the competitive performance.
\begin{table}[!]
\centering
\small
% \scalebox{0.9}{
\renewcommand\arraystretch{0.95}
\begin{tabular}{lcc}
\toprule
\multirow{1}{*}{Model} & \multicolumn{1}{c}{$K$} & \multicolumn{1}{c}{IWSLT14 DE$\rightarrow$EN} \\
\midrule
% NAT-base & 8.32\% & 7.10\%\\
GLAT & - & 32.49 \\
\midrule
\multicolumn{1}{l}{\multirow{3}{*}{RenewNAT (N=6)}}
& 2 & 33.00 \\
& 3 & 32.41 \\
& 4 & 32.03 \\
\midrule
\multicolumn{1}{l}{\multirow{3}{*}{RenewNAT (N=12)}}
& 4 & 32.23 \\
& 6 & 33.00 \\
& 8 & 32.90 \\
\bottomrule
\end{tabular}
\caption{Performances on IWSLT14 DE$\rightarrow$EN with different layer number of MLM module $K$.}
\label{tb.layer_effect}
\end{table}

\section{Ablation Study}
\noindent In this session, We mainly explore the influence of different factors for RenewNAT, including the corresponding layers to split the decoder, specific training strategy for MLM sub-module and the best decoding scheme. 
Besides, we also give more insights to show the effectiveness of RenewNAT compared with base models. 

% We mainly conduct experiments on IWSLT14 DE $\leftrightarrow$ EN and choose the base model GLAT.

% And we use the raw data of IWSLT14 DE $\leftrightarrow$ EN to evaluate how much RenewNAT can reduce dependence of training NAT models on AT models. Our experiments are mainly conducted on GLAT.

\subsection{Effect of Layers of MLM Sub-module}
Since RenewNAT splits the decoder into two sub-modules and adopts $N-K$ and $K$ decoder layers for NAT decoder and MLM sub-module respectively, we wonder how the performance evolves through the number of MLM sub-module layers $K$. Also, we extend the decoder layers to 12. As shown in Table~\ref{tb.layer_effect}, when $K$ is 2, RenewNAT performs best, but the performance degenerates quickly when $K$ is 4, That's because when $K$ is larger, the layers of NAT sub-module reduce, leading to bad quality of potential translation. Besides, it seems that adding the layers of each sub-module simply has no great improvements. However, with the given decoder layers $N$, the number of $K$ has a significant effect on the performance, requiring a good balance of two sub-modules.

% how to allocate the decoder layers effectively needs further exploration. Besides, we analyse the effects of adding the layers simply. As shown in Table~\ref{tb.layer_effect}, 

% and use \textit{K} and \textit{N-K} decoder blocks for the full NAT decoder module and MLM module respectively, we need to explore how the performance evolves through the number of MLM module layers \textit{K}. We set \textit{K} to be 2, 3, 4 respectively and compare them on IWSLT14 DE $\rightarrow$ EN. The results are shown in Table~\ref{tb.layer_effect}. We find that when \textit{K} is 2, RenewNAT performs best, but the performance degenerates significantly when \textit{K} is 4. That's because when \textit{K} is larger, the number of the fully NAT decoder module layers is smaller, causing the fully NAT decoder module is unable to performance well and MLM module is difficulty to repair the result.

\subsection{Effect of Training Strategy of MLM Sub-module}
We also experimented with different training strategies for MLM sub-module, the results are shown in Table~\ref{tb.stragies}. Since MLM sub-module is designed to derive useful target-side information, 
First, we adopt the masking operation to realize this. We set the second module to renew each word in potential translation completely and train the second module as CMLMC~\cite{huang2022improving}. As a result, the performance declines seriously.
Besides, from the perspective of MLM sub-module input, we propose several reasonable inputs with masking: target tokens, output tokens which are the output of NAT sub-module, and mixed tokens which mix ground-truth tokens and output tokens. Finally, taking mixed tokens as input outperforms others. In our opinion, mixed tokens can give some useful information about potential translation, which is matching to the inference process. 
Comparatively, output tokens may conclude many errors during training, which makes the training of MLM sub-module difficult and unstable.

% To evaluate the effect of different training strategies, we adopt the approaches from two aspects. Firstly, from the perspective of module, we change MLM sub-module into NAT sub-module. So it means that the second module needs to renew every word in the sentence completely. From the perspective of MLM module input, we summarize three data input: groundtruth tokens, the output of first module and mixed data which combine groundtruth tokens and the output of first module with a replacing ratio $\lambda$.

% Table~\ref{tb.stragies} presents the effect of different training strategies on IWSLT14 DE $\rightarrow$ EN. (describe in detail)

\begin{table}[!]
\centering
\small
% \scalebox{0.9}{
\begin{tabular}{lcc}
\toprule
\multirow{1}{*}{Model} & \multicolumn{1}{c}{IWSLT14 DE$\rightarrow$EN} \\
\midrule
% NAT-base & 8.32\% & 7.10\%\\
GLAT & 32.49 \\
\cmidrule{1-2}
RenewNAT (learning strategies) & \\
\ \ w/ renew [MASK] & 33.00 \\
\ \ w/ renew completely & 31.14 \\
\ \ w/ cmlmc & 32.73 \\
\cmidrule{1-2}
RenewNAT (inputs) & \\
\ \ w/ output tokens & 32.73 \\
\ \ w/ mixed tokens & 33.22 \\
\ \ w/ target tokens & 33.00 \\
\bottomrule
\end{tabular}
\caption{Performances on IWSLT14 DE$\rightarrow$EN with different training strategies based on GLAT.}
\label{tb.stragies}
\end{table}

% \subsection{Effect of Distinguish Strategy}
% \begin{figure}[ht]
% \centering
% \includegraphics[width=0.45\textwidth]{images/figure_confi.pdf}
% \caption{The distribution of confidence score for GLAT and Vanilla NAT on WMT14 DE-EN}
% \label{fig:confidence}
% \end{figure}

\subsection{Effect of Distinguish Strategy}
\begin{table}[!]
\centering
\small
\scalebox{0.8}{
\begin{tabular}{lcccccc}
\toprule
\multirow{2}{*}{Model} & \multicolumn{3}{c}{DE$\rightarrow$EN} & \multicolumn{3}{c}{EN$\rightarrow$RO}\\
& BLEU & Rep & Comet & BLEU & Rep & Comet \\
\midrule
% NAT-base & 8.32\% & 7.10\%\\
GLAT & 29.51 & 1.05\% & 0.27 & 31.65 & 1.09\% & 0.29\\
RenewNAT & 29.66 & 0.71\% & 0.30 & 32.52 & 0.59\% & 0.36\\
\ \ w/o MLM & 29.02 & 1.32\%& 0.24 & 32.10 & 1.06\% & 0.31\\
\bottomrule
\end{tabular}
}
\caption{BLEU score and token repetition ratio on WMT14 DE$\rightarrow$EN and
WMT16 EN$\rightarrow$RO.}
\label{tb.mlmworks}
\end{table}

\begin{figure}[ht]
\centering
\includegraphics[width=0.3\textwidth]{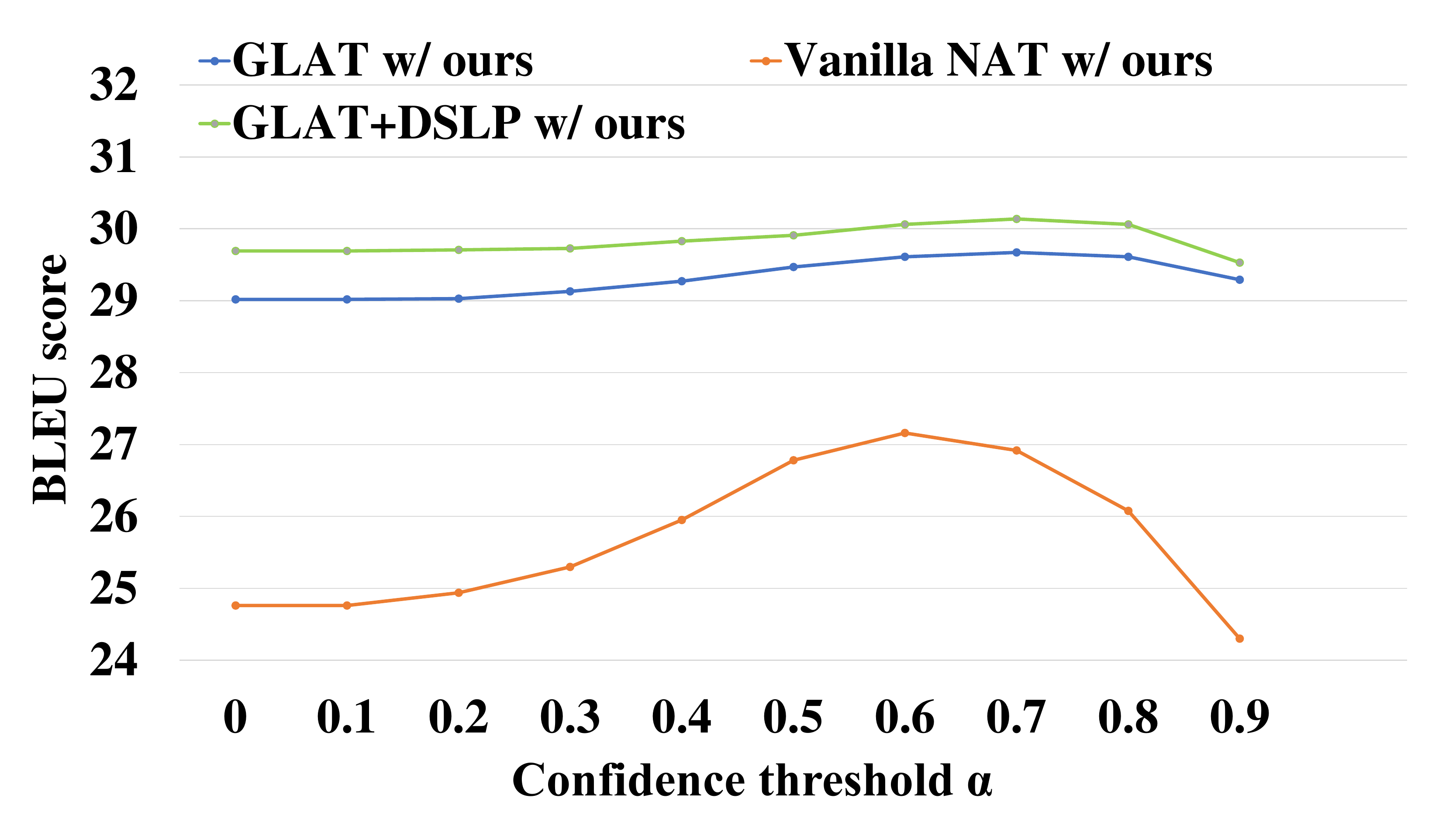}
\caption{BLEU score with different confidence threshold on WMT14 DE$\rightarrow$EN.}
\label{fig:confidence}
\end{figure}

As mentioned above, there always exist some errors in potential translation, and these tokens hurt the second sub-module to capture the target side dependency. We adopt the masking operation to discard these incorrect tokens, as a result, 
% since there are no ground truth tokens during inference, 
how to effectively distinguish them is critical for the performance of our RenewNAT, we try some simple methods to achieve this purpose. 
% 表格，random，依据confidence
Firstly, we give a fixed masking ratio $\delta$ to mask the tokens with the lowest confidence, for the assumption that there is an almost fixed ratio of errors in potential translation, we find this method can obtain the high performance when $\delta$ is in [0.2, 0.5].
However, it is obvious that different potential translation results may have different incorrect ratios and the fixed ratio seems not optimal. Thus we 
follow another simple and effective method which utilizes a confidence threshold $\alpha$ to choose the incorrect tokens.
% During inference, RenewNAT firstly generates a potential translation result based on the NAT decoder sub-module, then masks the incorrect tokens in the potential translation result and adopts the MLM decoder sub-module to renew them. 
% Therefore, masking the incorrect tokens rightly is very important. First of all, we adopt a fixed masking ratio $\delta$ to mask the tokens with the lowest confidence. We find that this approach can obtain the high performance when $\delta$ is in [0.2, 0.5]. However, different potential translation results may have different incorrect ratios. Finally, we follow a simple and effective method which utilizes a confidence threshold $\alpha$ to choose the incorrect tokens. 
Specifically, when the confidence of target tokens is lower than $\alpha$, we regard them as incorrect tokens and mask them. 
Moreover, the best $\alpha$ is related to the base model, which is also easy to understand as the quality and distribution of the potential translation generated by different models are truly inconsistent.
As shown in Figure~\ref{fig:confidence}, on WMT14 DE$\rightarrow$EN dataset, Vanilla NAT and GLAT achieve the best performance when $\alpha$ is set to 0.6 and 0.7 respectively. 
More exploration of the reasons and the best threshold $\alpha$ for different settings are presented in Appendix.
% As shown in Figure~\ref{fig:confidence}, we suppose that Vanilla NAT generates more tokens with low confidence compared with GLAT, so we need to reduce the threshold $\alpha$ for Vanilla NAT to mask less target tokens and provide MLM sub-module with enough target side information. 
% More best threshold $\alpha$ on different base models will be presented in Appendix.

\subsection{Performance of MLM Sub-module }
RenewNAT splits the traditional decoder module into two sub-modules, called NAT sub-module and MLM sub-module. Need to notice that NAT sub-module is similar to the traditional NAT decoder, just with fewer layers. Our assumption is that MLM sub-module can derive useful target side information from NAT sub-module. To give more insights to verify this, we wonder if MLM sub-module truly improves the performance and how this works. As shown in Table~\ref{tb.mlmworks}, we compare our RenewNAT with the corresponding base model GLAT and the one without MLM sub-module from different aspects. We can observe that RenewNAT achieves great improvements on BLEU score and Comet score of MLM sub-module (29.66 vs. 29.02, 32.52 vs. 32.10, 0.30 vs. 0.24 and 0.36 vs. 0.31), besides, MLM sub-module can effectively alleviate the repetition problem.
The comparison between the base model further verifies the effectiveness of our RenewNAT, certificating that MLM sub-module can successfully capture the internal dependency of target tokens, then predicts the masked tokens effectively.

% We focus on MLM sub-module and analyze the performance of it. Table~\ref{tb.repetition} shows that NAT sub-module is able to generate the comparable translation results with the base models, and MLM sub-module improves the performance with + BLEU score. It certificates that MLM sub-module successfully captures the target side information, and then predicts rightly the masked tokens.

\subsection{Effect on Raw Data}
Since we all conduct experiments on distilled data above, we wonder the effect of our RenewNAT on raw data, we conduct experiments on IWSLT14 DE $\leftrightarrow$ EN datasets without distillation.
% To evaluate how much RenewNAT can reduce dependence of training NAT models on AT models, we compare the performance of GLAT w/ and w/o RenewNAT on the raw data of IWSLT14 DE $\leftrightarrow$ EN. 
The results are shown in Table~\ref{tb.raw_data}. As we can see, RenewNAT outperforms the corresponding base model on IWSLT DE$\rightarrow$EN with + 1.42 BLEU score,
% with the help of our framework, GLAT is significantly improved on raw data with +1.42 BLEU score gain which is able to 
demonstrating the robustness of our framework. 
% Therefore, RenewNAT is actually able to improve the ability of base models and reduce the dependency on distill data.
\begin{table}[!]
\centering
\small
% \scalebox{0.9}{
\begin{tabular}{lcccc}
\toprule
\multirow{2}{*}{Model} & \multicolumn{2}{c}{EN$\rightarrow$DE} & \multicolumn{2}{c}{DE$\rightarrow$EN}\\
& BLEU& Comet& BLEU & Comet\\
\midrule
% NAT-base & 8.32\% & 7.10\%\\
GLAT &  22.37 & -0.19 & 28.55 & 0.08 \\
w/ RenewNAT & 22.72 & -0.20 & 30.12 & 0.14 \\
\bottomrule
\end{tabular}
\caption{Performances on the raw IWSLT14 EN$\leftrightarrow$DE.}
\label{tb.raw_data}
\end{table}

\subsection{Effect of Source Input Length}

\begin{figure}[!]
\centering
\includegraphics[width=0.3\textwidth]{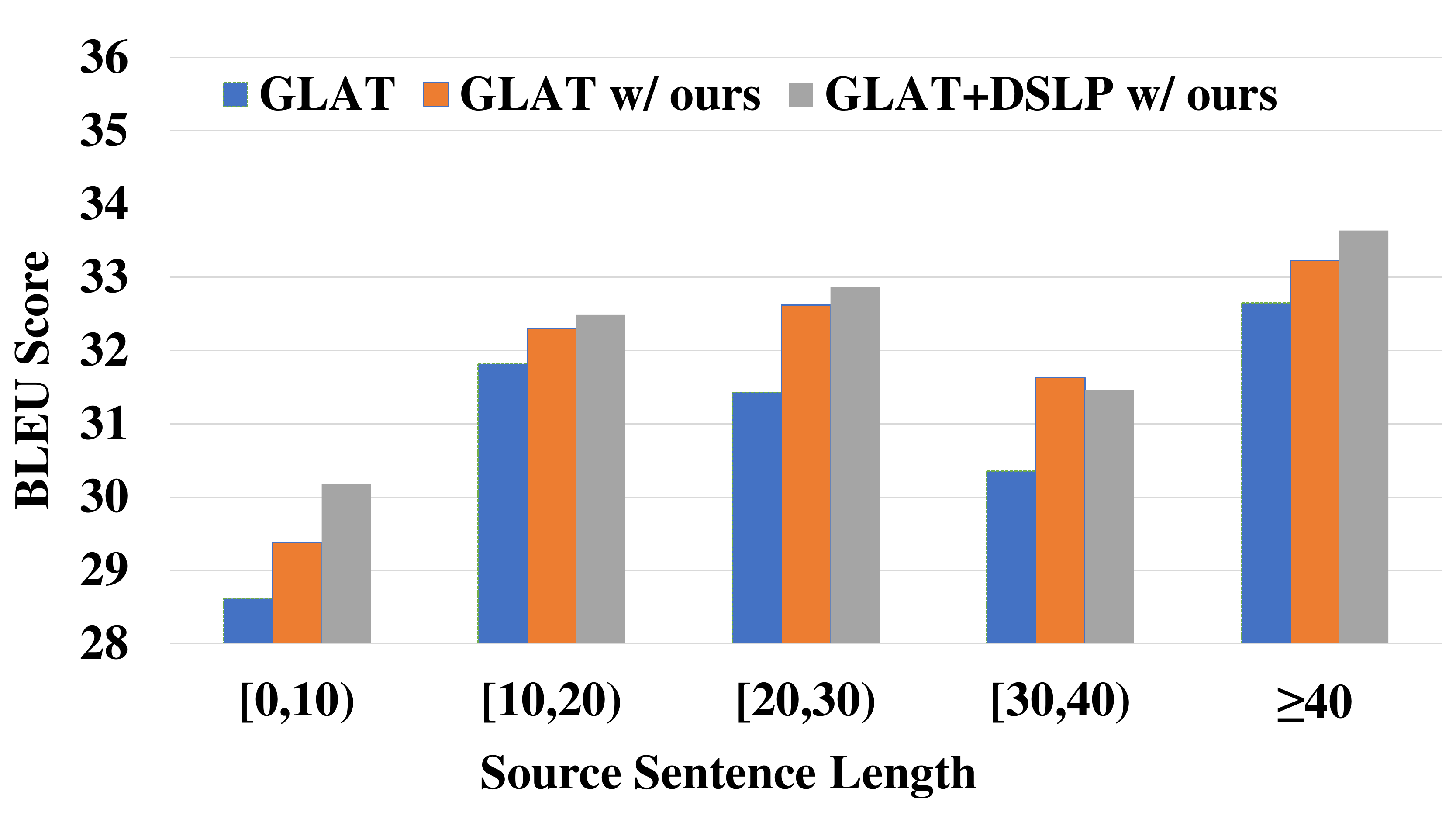}
\caption{BLEU score with different source sentence length on WMT16 EN$\rightarrow$RO.}
\label{fig:source_length}
\end{figure}
We also explore the effect of sentence length for our RenewNAT. Specifically, we divide the source sentences of the WMT16 EN-RO test set into 5 intervals by the length after BPE operation and compute the
BLEU score for each interval. Figure~\ref{fig:source_length} shows the histogram results. We find that RenewNAT outperforms the correlated base model in each interval.
When the source sentence length is in [20,40], RenewNAT achieves great improvement.
As a result, RenewNAT can boost the performance under different length scenes. 

\section{Conclusion}
% In this work, we proposed RenewNAT, a simple yet effective framework with high effecency and effectiveness, which first generate the potential translation to provide useful target side information and then renew them and learn to capture the internal dependency of target tokens. We conduct experiments on five translation tasks with several base models (vanilla NAT and two strong fully NAT models). Results demonstrate that RenewNAT effectively improve the translation with little effect on decoding latency. In future works, we will explore more base models and whether our proposed framework can be introduced to iterative NAT models to reduce decoding expense to achieve comparable performance.
In this work, we proposed RenewNAT, a simple yet effective framework with high efficiency and effectiveness. 
It first generates the potential translation to provide useful target side information and then renews them to capture the internal dependency of target tokens. 
We conduct experiments on five translation tasks with multiple representative base models (vanilla NAT and two strong fully NAT models). 
The results show that RenewNAT effectively improves the translation with little effect on decoding latency. 
In the future, we will explore more base models and whether our framework can be used in iterative NAT models to reduce decoding expense and meanwhile maintain performance advantage.
\section{Acknowledgements}
We would like to thank the anonymous reviewers for their constructive comments. This work is supported by the National Science Foundation of China (NSFC No. 62206194), the Natural Science Foundation of Jiangsu Province, China (Grant No. BK20220488), and the Project Funded by the Priority Academic Program Development of Jiangsu Higher Education Institutions.
This work is also supported by Beijing Academy of Artificial Intelligence (BAAI).

\bibliography{aaai23}
\end{document}